\documentclass[sigconf]{acmart}

\AtBeginDocument{%
  \providecommand\BibTeX{{%
    \normalfont B\kern-0.5em{\scshape i\kern-0.25em b}\kern-0.8em\TeX}}}

\setcopyright{acmcopyright}

\copyrightyear{2019} 
\acmYear{2019} 
\acmConference[DLP-KDD'19]{1st International Workshop on Deep Learning Practice for High-Dimensional Sparse Data}{August 5, 2019}{Anchorage, AK, USA} 
\acmBooktitle{1st International Workshop on Deep Learning Practice for High-Dimensional Sparse Data (DLP-KDD' ), August 5, 2019, Anchorage, AK, USA}
\acmPrice{15.00}
\acmDOI{10.1145/3326937.3341253}
\acmISBN{978-1-4503-6783-7/19/08}

\begin{document}

\title{An Adaptive Approach for Anomaly Detector Selection and Fine-Tuning in Time Series}

\author{Hui Ye}
\affiliation{%
  \institution{Alibaba Inc}
  \city{Beijing}
  \country{China}
}
\email{yehui.yh@alibaba-inc.com}

\author{Xiaopeng Ma}
\affiliation{%
  \institution{Alibaba Inc}
  \city{Beijing}
  \country{China}
}
\email{xiaopeng.mxp@alibaba-inc.com}

\author{Qingfeng Pan}
\affiliation{%
  \institution{Alibaba Inc}
  \city{Beijing}
  \country{China}
}
\email{qingfeng.pqf@alibaba-inc.com}

\author{Huaqiang Fang}
\affiliation{%
  \institution{Alibaba Inc}
  \city{Beijing}
  \country{China}
}
\email{huaqiang.fhq@alibaba-inc.com}

\author{Hang Xiang}
\affiliation{%
  \institution{Alibaba Inc}
  \city{Beijing}
  \country{China}
}
\email{xingzhi.xh@alibaba-inc.com}

\author{Tongzhen Shao}
\affiliation{%
  \institution{Alibaba Inc}
  \city{Beijing}
  \country{China}
}
\email{yeqing.stz@taobao.com}




\begin{abstract}
The anomaly detection of time series is a hotspot of time series data mining. The own characteristics of different anomaly detectors determine the abnormal data that they are good at. There is no detector can be optimizing in all types of anomalies. Moreover, it still has difficulties in industrial production due to problems such as a single detector can't be optimized at different time windows of the same time series. This paper proposes an adaptive model based on time series characteristics and  selecting appropriate detector and run-time parameters for anomaly detection, which is called ATSDLN(Adaptive Time Series Detector Learning Network). We take the time series as the input of the model, and learn the time series representation through FCN. In order to realize the adaptive selection of detectors and run-time parameters according to the input time series, the outputs of FCN are the inputs of two sub-networks: the detector selection network and the run-time parameters selection network. In addition, the way that the variable layer width design of the parameter selection sub-network and the introduction of transfer learning make the model be with more expandability. Through experiments, it is found that ATSDLN can select appropriate anomaly detector and run-time parameters, and have strong expandability, which can quickly transfer. We investigate the performance of ATSDLN in public data sets, our methods outperform other methods in most cases with higher effect and better adaptation. We also show experimental results on public data sets to demonstrate how model structure and transfer learning affect the effectiveness.
\end{abstract}



\maketitle

\section{Introduction}
\par
Internet-based services have strict requirements for continuous monitoring and in-time anomaly detection, Specifically, monitoring performance ability and detecting performance anomalies are important. Such as, e-commerce platforms need to monitor income index and broadcast alert when obvious income decrease happens. 
\par
From the perspective of data science, key performance indexes are usually portrayed as time series, and potential faults in application are portrayed as anomaly. An anomaly (An outlier) in time series, is a data point or a group of data points which significantly different from the rest of the data points\cite{laptev2015generic}. Due to the large amounts of performance indexes and anomalies, human monitoring of these indexes is impracticable which leads the demand for automated anomaly detection using Machine Learning and Data Mining techniques\cite{malhotra2016lstm, shipmon2017time, hundman2018detecting, xu2018unsupervised}. Many fast and effective anomaly detectors were designed to localize these anomalies\cite{chandola2009anomaly}, such as outlier detector\cite{breunig2000lof}, change point detector\cite{kawahara2007change}. 
Although anomaly detectors have proven effective in certain scenarios, applying them to internet-based services remains a great challenge\cite{liu2015opprentice}. Due to the large-scale distributed monitoring vision and complex trends of indicators, it's almost impossible to detect anomalies in all scenarios with one type of detector. In order to ensure the performance of the anomaly detection approach, expertise-based rules are required for detector selection and run-time parameters fine-tuning\cite{liu2015opprentice}. Furthermore, when a detector system is deployed online, the run-time parameters of anomaly detector are usually required to adjust according to real-time changes.
\par
It's hard to propose one general approach to detect all types of anomaly, such as significant decrease or increase can be detected by static threshold directly, continuous minor changing can be detected by change point detector more quickly. State-of-the-art detectors are usually designed to detect one type of anomaly\cite{laptev2015generic}. When the multi-detector detection result voting method is adopted, each detect needs to traverse all detectors and candidate run-time parameters combinations. The effect is greatly influenced by the data set and voting rules and it is very time-consuming, which do not meet the demands of industrial real-time monitoring scenarios. Our proposed framework named ATSDLN, tackles the above challenges through an adaptive time series anomaly detector learning network.

\section{Methods}
\begin{figure*}
\centering
\includegraphics[width=13cm]{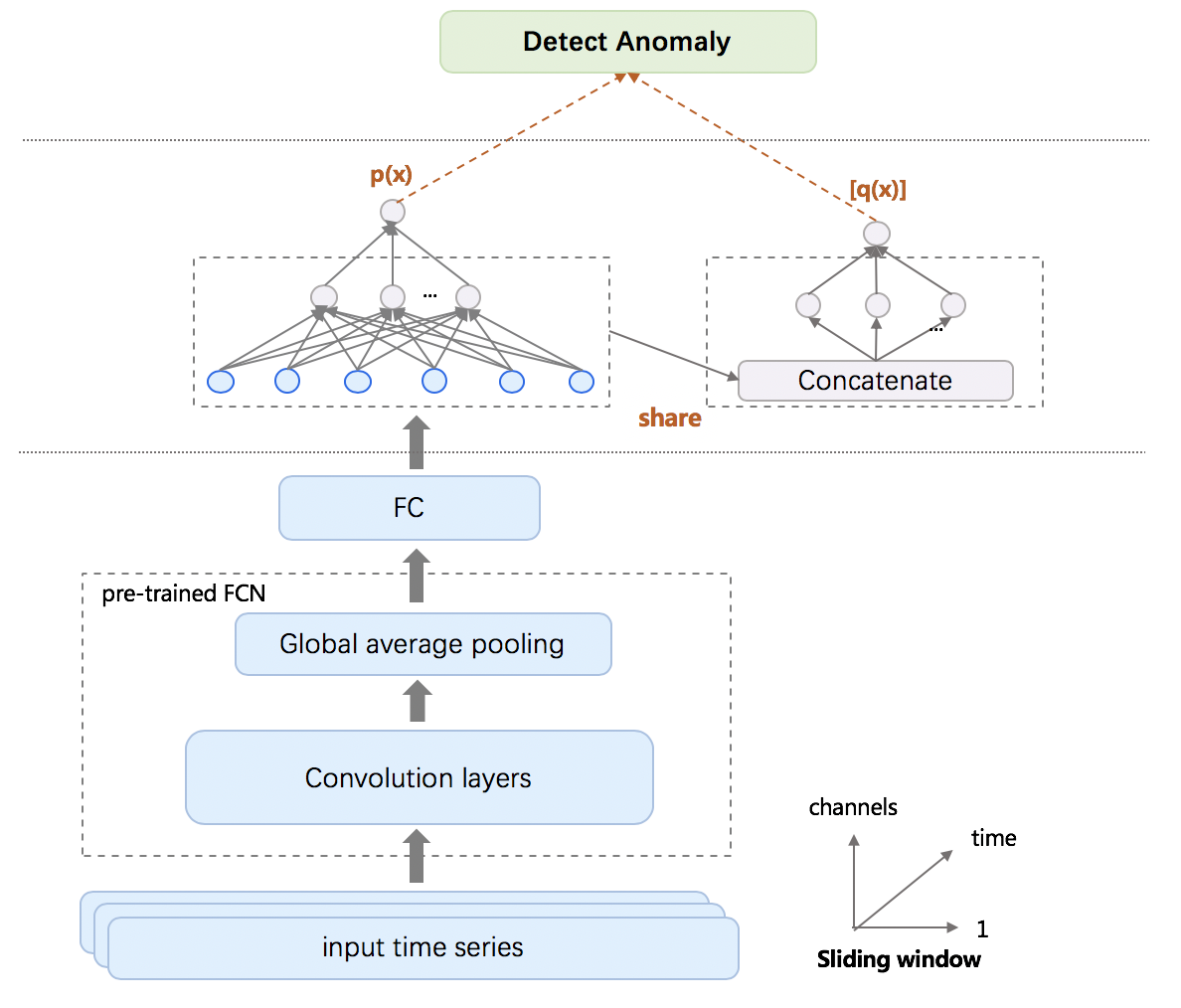}
\caption{Whole Net Structure, left represents anomaly detectors classification task, right represents run-time parameters fine-tuning task. Blue layers are shared by the two sub-networks.}
\label{fig:fcnmultinet}
\end{figure*}
\par
Under the background of large industrial data scale, complicated index system and an unusually large variety, on the one hand, time series data usually changes with business changes. The same time sequence may have great differences in different stages of business projects; on the other hand, influenced by commercial data and users' behaviors, there are different low ebbs of the peak flow on holidays, daytime and nights, big promotions and so on, which cause the natural differences in data. If we do not consider self-adaption when doing anomaly detection, we cannot balance between the false positive rate and the false negative rate. Therefore, choosing a universal detector to adapt to all data and scenarios is unworkable. Multi-detection algorithm fusion is a very effective method to improve the time series anomaly detection field, which is usually conducted in the two stages as follows:
\par
\verb|The anomaly detection stage|: it is realized by selecting the appropriate detector for the time series of the input.
\par
\verb|The alarm convergence stage|: it is realized by using the abnormality that is detected by each detector as the input. The alarm convergence can be achieved with the method of voting or time series feature modeling.
\begin{itemize}
\item {\verb|Voting method|}: absolute majority vote, relative majority vote, weighted vote, etc.
\item{\verb|Deep learning|}: time series modeling of the detected anomalies.
\end{itemize}
\par
Both of them are of highly expandability and support dynamic expansion of anomaly detectors. The former is self-adaption based on the original time series of the input, which is more flexible, this study takes the former. As is shown in the experimental chapter, the single detector is lower than our model in term of the accuracy, recall, and f1, and the error rate is relatively high. The starting point of this study is to set a certain sliding window size for the time series, and optimize the accuracy, recall and false positive rate of the anomaly detection through using the detector and run-time parameters for the self-adaption selection of the current sliding window time series.
\par
Since different detectors have their own characteristics which determine the type of time series they are good at, it is natural to think about to determine which the detectors and run-time parameters are suitable for by the features of the time series. We call this way the manual rule maintenance detector and run-time parameters selection. The core work is to determine what features of time series and what threshold should be used for judgment (for instance, non-stationary time series with long-term trends can adopt dynamic thresholds). The advantage of such artificial rules is that it has strong interpretability. However, it is true that the determination of these rules relies on manual experience, which is difficult to enumerate the rules. As the data accumulation rules become more and more difficult to maintain, the abnormal coverage, correctness, versatility and expandability of the rules are also great challenges.
\par
Fortunately, in the era of artificial intelligence, it is natural to think of using models to replace labor. Generally speaking, time series classification using traditional machine learning methods (such as KNN, DTW) can achieve better results. However, as for big data, deep learning tends to defeat traditional methods. Until recently, a paper relevant to the research was published by Fawaz H I et al. \cite{fawaz2018transfer}, which has demonstrated the feasibility of transfer learning method for different time series data. The author argues that FCN can learn time series representation well when the amount of data is sufficient, and believes that the features extracted by the deep network for the time series data are as similar and inherited as CNN in terms of time series. Moreover, one of the challenges for supervised learning is the large number of labeled data. Unfortunately, it is not readily available for the real-world labeled data problem tended to the high cost and longtime consuming. This problem in essence involves using transfer learning to obtain a solution. It can be seen that the solution based on the transfer learning becomes a better choice for the self-adaption anomaly detection problem.
\par
\textbf{A new ATSDLN model} is proposed in the paper, which realized an adaptional classification of time series anomaly detectors and run-time parameters selection by combining transfer learning and dynamic adaptive joint learning. It is a pre-trained model based on public data sets for transfer learning. Figure \ref{fig:fcnmultinet} is our frame diagram. The model supports multiple channels, and can input the original time series, prediction time series or residual sequence.  
\par
From the bottom to the top, the first part is the Fully Convolutional Neural Network (FCN), which is made up of Convolution layers and Global average pooling layer. As the Figure \ref{fig:fcnmultinet} shows, transfer learning is applied to the FCN layer and fine-tuning in the FC layers, which makes the network parameters initialized better, so as to speed up the training and convergence and improve the performance of time series classification model. The main function of this part is to learn the rich time series representation by means of a large amount of training data, and then produce time series representation. This part introduces the ability of the migration learning enhancement model to extract the ability of time series representation, to deal with the problem of marking sample sparseness and model mobility.
\par
The second part is composed of two sub-networks, both of which are supervised classification models. The left part is responsible for the classification of the detector, while the right part is responsible for the classification of the corresponding run-time parameters of the detector, the two parts can jointly study. The expression learned through the detector classification task will be used as the input of the run-time parameters selection task, which can assist the learning of the run-time parameters. Both of the sub-networks have the problems of supervised classification. From the figure \ref{fig:fcnmultinet}, it can be seen that the output of p(x) determines a certain detector uniquely for the current time series, and [q(x)] is the run-time parameters that the current time series and anomaly detector choose.
Because the size of the candidate run-time parameters sets of each detector is inconsistent, the last layer width of the right network follows the left as the side detector changes, that the model supports flexible addition and deletion detectors. It can be known from the above that the selection of the run-time parameters on the right side depends not only on the time series representation, but also on the detector selected by the network on the left side. So in this part, the expression learned in the left detector classification task is shared to the task of the right run-time parameters selection on the right and is taken as its input to assist in learning.
\par
The third part, which is on the top, is the execution module for the anomaly detection. It detects the anomaly of the detector and the run-time parameters which is selected when time series use models.
\par

\section{EXPERIMENTAL SETTING}
\par
The following parts form the core components of an joint learning approach. The two sub-networks in our approach refers to anomaly detectors classification task and run-time parameters fine-tuning task, which means the network predicts optimal detector and fine-tunes the run-time parameters simultaneously without human interfering. Firstly, we collect some classical detectors, which were proposed to detect anomaly in different context. Secondly, a new evaluation criterion was proposed to evaluate the performance of these detectors in each time series data, this process also generates the label of our two sub-tasks. Thirdly, an adaptive model is trained to extract deep features of time series, which is crucial for optimal detector prediction and the run-time parameters fine-tuning tasks. Lastly, we transfer this representation learned from public data sets to other unseen data sets and evaluate the usability of transfer learning in time series anomaly detection.

\subsection{Datasets}
We set the different sizes of sliding window on webscope S5 data sets\footnote[1]{https://research.yahoo.com/} for the experimental sample, which contains outliers and change points, and use the UCR Time series Classification Archive\footnote[2]{https://www.cs.ucr.edu/} as the source data sets for transfer learning.

\textbf{Webscope S5} is a labeled anomaly detection data set. There are 367 time series in the data sets, each of which contains between 741 and 1680 data points at regular intervals. Each time series is accompanied by an indicator series with 1 if the observation was an anomaly, and 0 otherwise. 
\textbf{UCR} is a time series classification data sets. There are 128 data sets with different applications. The classification type of these data sets is from 2 to 60, and the the size of data sets is from 20 to 8926.
\par
Through traversing the candidate detector and the combining operational parameters, the optimal detector and run-time parameters are selected for the time series as a training data for supervised learning. Then, by carrying out the pre-training of the transfer learning on the UCR time series classification data set, the data volume problem of the training data charged by the meter is solved.
\subsection{Evaluation criterion}
\begin{table*}
  \caption{Evaluation Metrics}
  \label{tab:Metrics}
  \begin{tabular}{ll}
    \toprule
    Metric &Description \\
    \midrule
    Precision & Precision is defined as TP/(TP + FP)  \\
    Recall & True positive rate or recall is defined as TP/(TP+FN) \\
    False positive rate & The false positive rate is defined as FP/(FP+TN) \\
    F1-score & F1-score is dened as 2 * precision * recall /(precision + recall) \\
    Error & Error is defined as FP/(TP+FP+FN) \\
    \bottomrule
  \end{tabular}
\end{table*}
\par
Experiments results were evaluated by comparing observed anomalies to true anomalies. In table \ref{tab:Metrics}, we present the evaluation measures of the model's such as precision and recall, Error which were used. FP denotes the number of false positive, FN the number of false negative, TP the number of true positive and TN the number of true negative.

\par
Number of true positive whose proportion in anomaly detection is small, in addition without considering precision's inability to accurately express the level of false positive ratio (or false alarm ratio), especially when true positive is zero, precision is always zero. there are not very good measures for assessing anomaly detection methods. In our situation, the high false positive ratio will cause alarm fatigue of the relevant personnel, which will lead to the decrease of the attention of monitoring alarm. However, the number of true negative is large, so the false positive rate is not sensitively enough as it grows very slowly. Therefore, we propose a new metric named Error which is defined as FP/(TP+FP+FN). 
\subsection{Detectors for time series anomaly detection}
\begin{figure}[h]
  \centering
  \includegraphics[width=\linewidth]{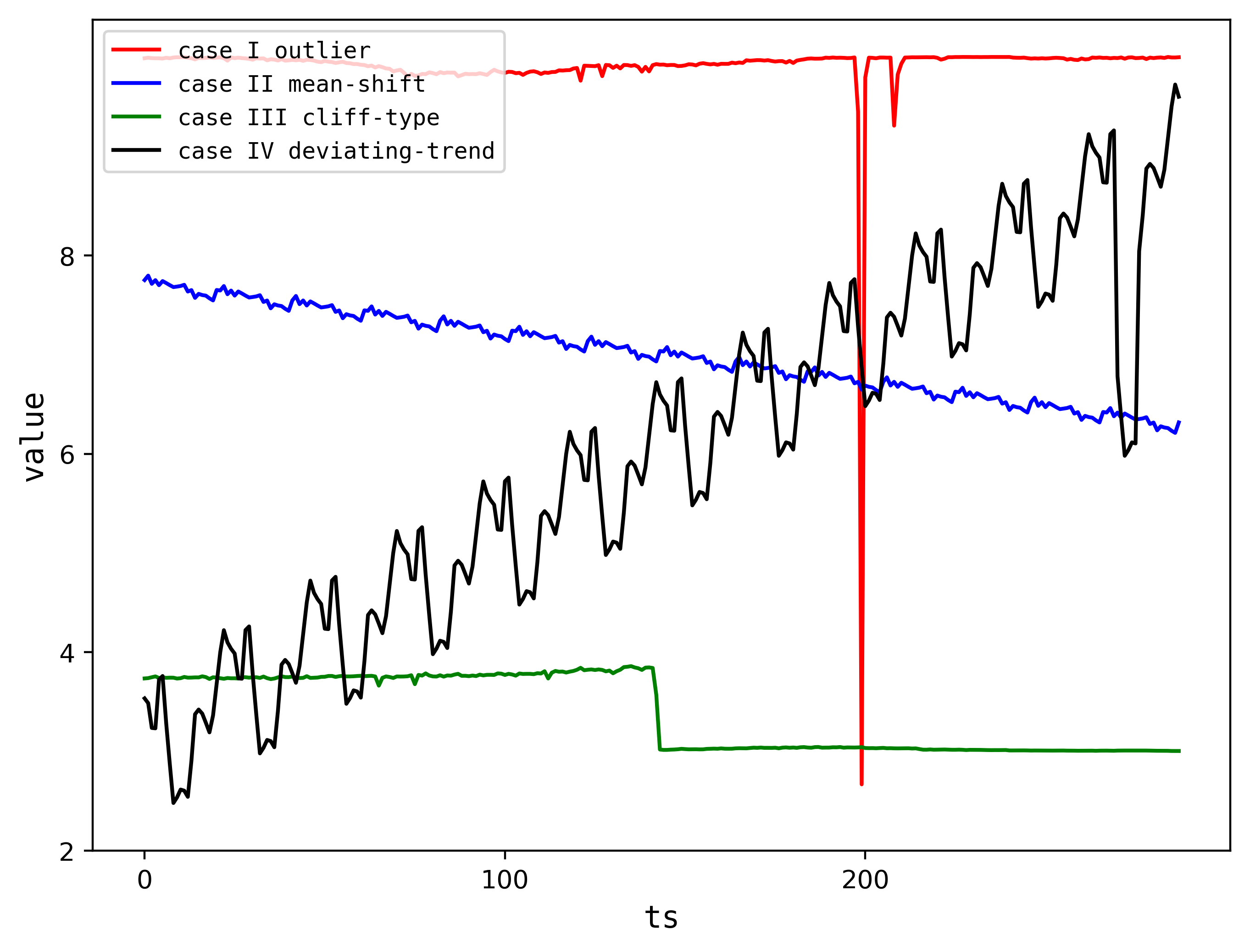}
  \caption{Example of anomaly types.}
\label{fig:anomaly classify}
\end{figure}

 According to the shape and context of time series anomaly, it can be summarized as outlier, mean-shift, cliff-type, deviating-trend, new-shape. See the table \ref{fig:anomaly classify} for details. The anomaly detectors used in ATSDLN are just the same as EGADS.
 
 \begin{table*}
  \caption{Anomaly and Detectors}
  \label{tab:my_label}
  \begin{tabular}{lll}
    \toprule
    type &descriptions &detector \\
    \midrule
    outlier & significantly different & KSigma/DBScan/LOF/Extreme LowDensity\cite{laptev2015generic,breunig2000lof}  \\
    mean-shift & sustained inapparent deviation & CUSUM changepoint\cite{kawahara2007change} \\
    cliff-type & switch to another sustained value & KernelDensity changepoint/KSigma/SimpleThreshold \\
    deviating-trend & not in line with fitting trend & STL decomposition\cite{cleveland1990stl} \\
    new-shape & un-similar with others & DTW similarity\cite{fu2011review} \\
    \bottomrule
  \end{tabular}
\end{table*}
\par
In addition, the parameters fine-tuning is as important as the accuracy of selecting the most suitable detector. Detector parameters are divided into two categories: the first is the common parameters needed by all detectors, including sliding window size, sensitivity, number of historical samples. The second is the internal parameters required by each detector algorithm, such as K-Multiple variance of KSigma, eps and minPts of DBScan, confidence and drift range of ChangePoint, search radius of DTW similarity, etc.
\section{RESULTS AND DISCUSSIONS}
\begin{figure}[h]
  \centering
  \includegraphics[width=\linewidth]{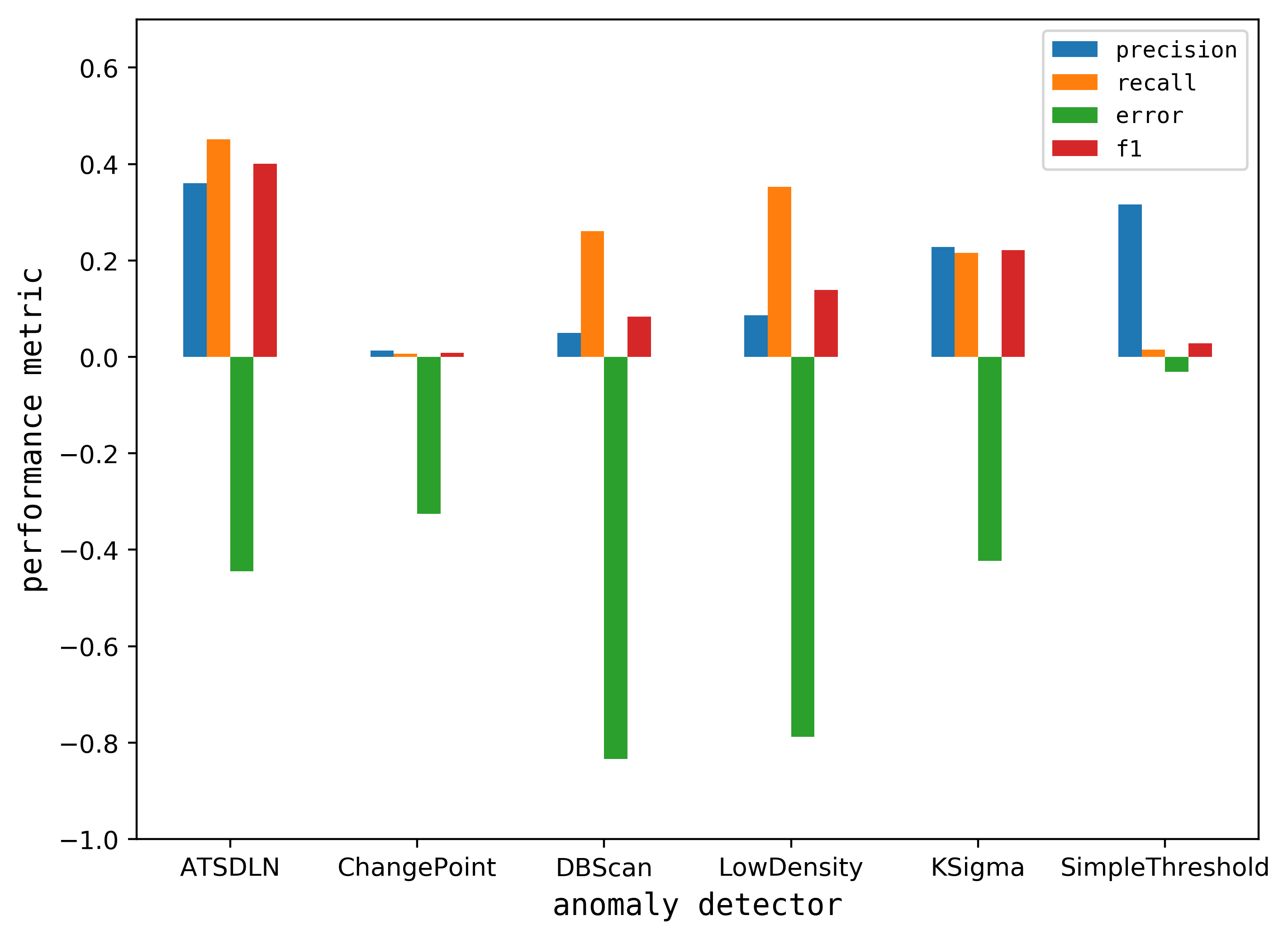}
  \caption{Anomaly model performance on different detectors(adaptively select best run-time parameters).}
\label{fig:expfig5}
\end{figure}
\par
The second chapter mentions that the network is composed of two sub-networks, both of which are supervised classification models. The output of p(x) determines a certain detector uniquely, and q(x) is the run-time parameters corresponding to the detector. With the determined detector and run-time parameters, it is possible to judge the abnormality of the time series execution abnormality detection. The evaluation of part of the experimental effect adopts the precision, recall and error described in evaluation criterion in Chapter 3.2.
\par
\begin{figure}[h]
  \centering
  \includegraphics[width=\linewidth]{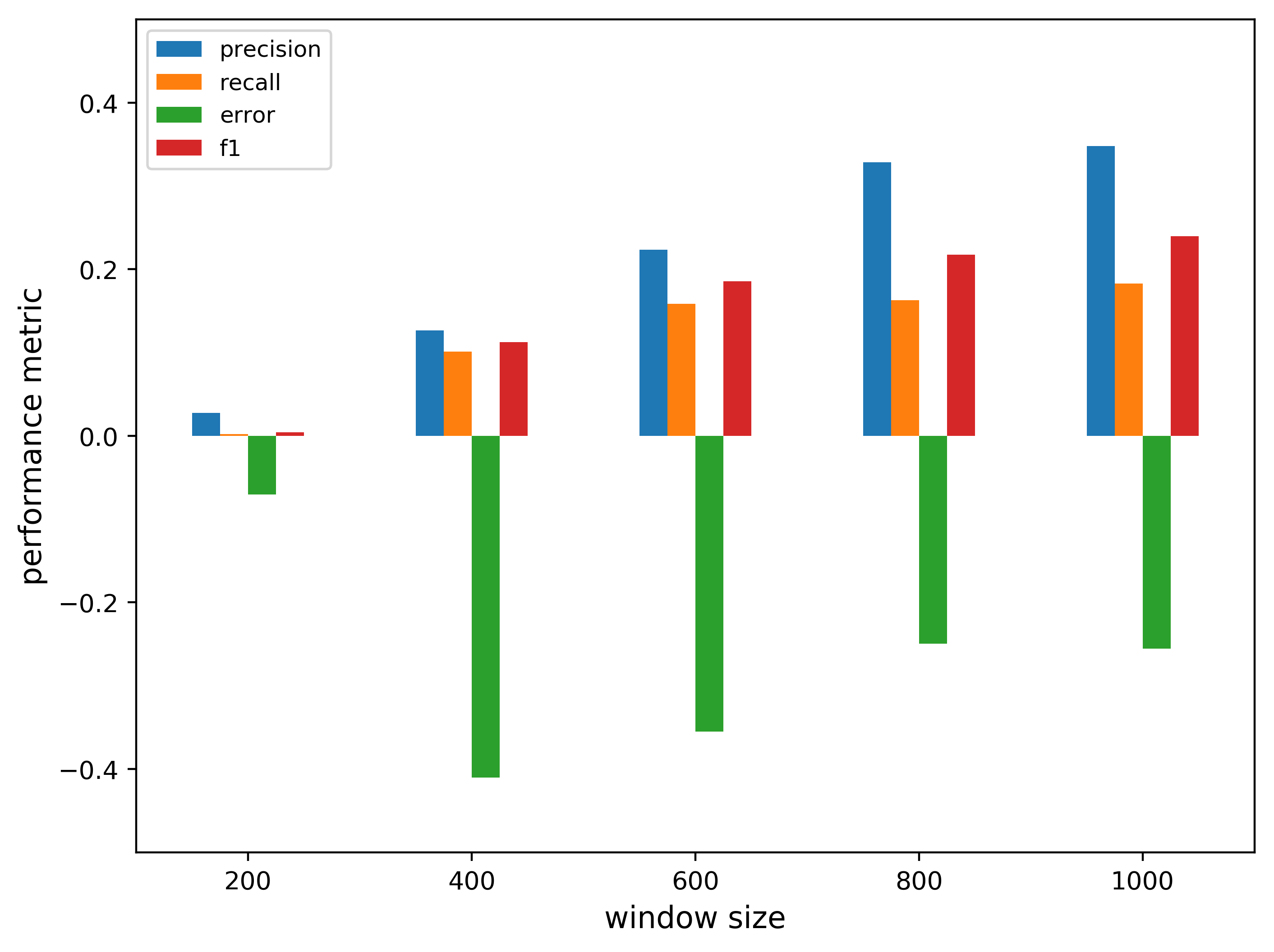}
  \caption{Baseline performance on different window size.}
\label{fig:baseline_windowsize}
\end{figure}

The main work of this paper is to select the appropriate detector as well as run-time parameters for a certain time series. The length of the time series is called the window size of the time series. The size of the window not only has relationship to the business attributes, but also influences the sensitivity of the detector's self-adaption selection. In theory, the smaller the window, the more sensitive the changes in the detector and parameters. The traditional voting method relies more on the accumulation of time series data and has poor adaptability. As is shown in Fig. \ref{fig:baseline_windowsize}, the horizontal axis is the window size of the time series, while the vertical axis if the evaluation index calculated by the abnormality detection result. It can be seen that the smaller the window, the worse the baseline effect. According to our experiments, the window size will not affect the performance of our model. The ATSDLN can better adapt to different window size. In order to compare the performance of different experiments, we choose the window size with 200 points. 

\subsection{Anomaly model performance analysis}
\par
\begin{figure*}[h]
  \centering
  \includegraphics[width=14cm]{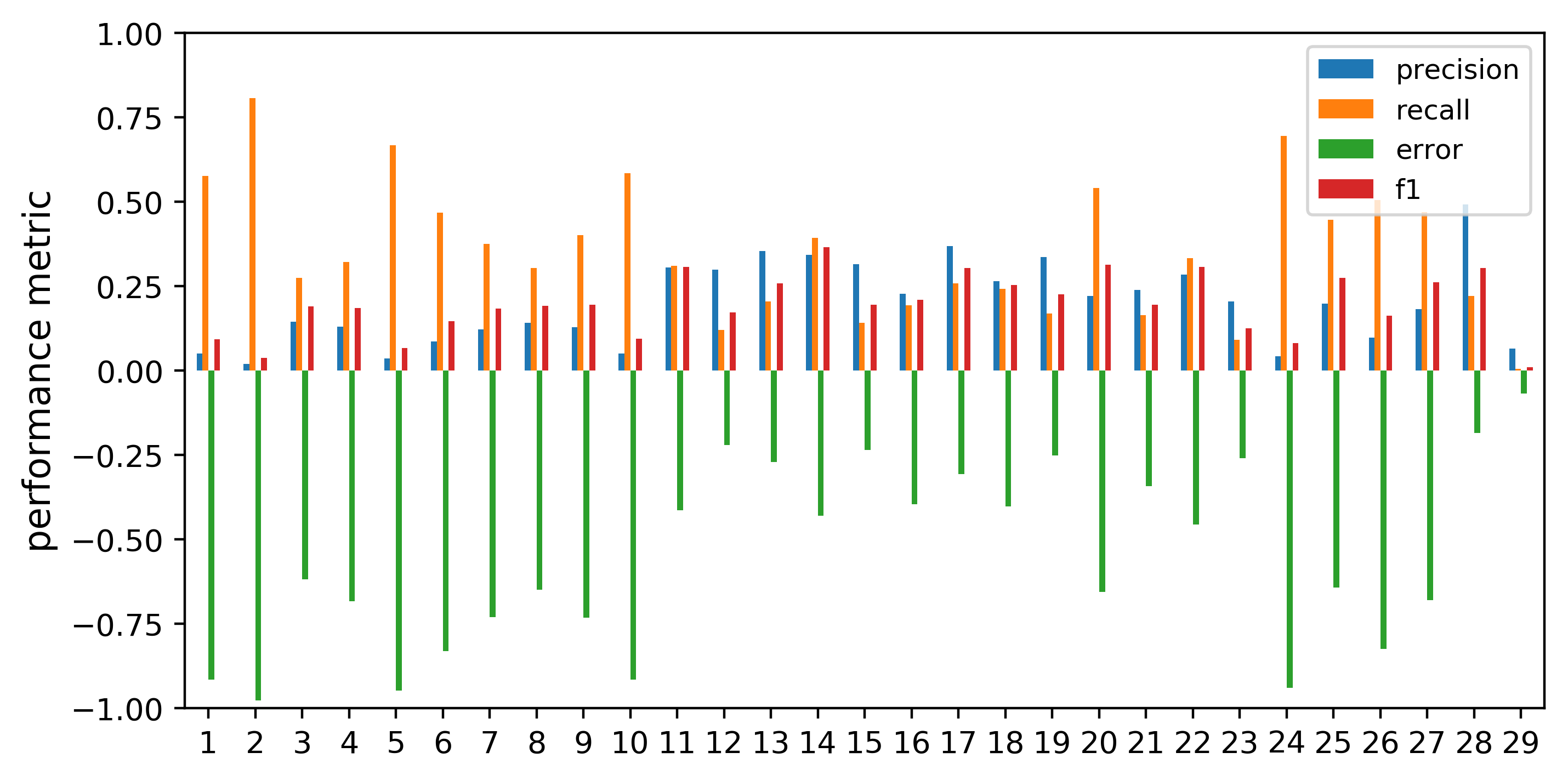}
  \caption{Anomaly model performance on different parameters.}
\label{fig:expfig4}
\end{figure*}

In order to explore the necessity for self-adaption selection detectors and operating parameters, 29 combination parameters of five detectors are selected which described in detectors for time series anomaly detection in Chapter 3.3. The experiments are performed on the yahoo public data set. The results are shown in Figure \ref{fig:expfig4}, the horizontal axis shows the 29 combinations of detectors and its parameters, the vertical axis shows the performance under each combination. There is no existence of fixed detector and parameters which can be optimal at the whole time series. In addition, the different effects of the run-time parameters are widely divergent under the situation when the detector is determined.
\subsection{Compare with single model}
\par
Figure \ref{fig:expfig5} compared the results of ATSDLN with other single detector models. It shows that the performance of our method is the best. The method proposed in the paper, regardless of accuracy, recall rate or F1, is superior to the single detector adaptive selection of optimal parameters, and the false positive rate is also reduced.
\subsection{Comparison of different network architectures}
In this paper, we compare several network architectures to investigate our proposed model's effectiveness. The controlled models are described as follows:
\begin{itemize}
\item {\verb|Baseline|}: The majority voting algorithm based on EGADS.
\item {\verb|LSTM-DNN|}: A hybrid neural network composed of Long short-term memory LSTM and DNN network.
\item {\verb|FCN-LSTM-DNN|}: The model adds a CNN layer to capture features based on the LSTM-DNN model.
\end{itemize}

\begin{table}
  \caption{Peformance of different network
architectures}
  \label{tab:t3}
  \begin{tabular}{lllll}
    \toprule
    \textbf{model type}    & \textbf{Precision} & \textbf{Recall} &\textbf{Error}&\textbf{F1} \\
    \midrule
    Baseline  & 0.0278  & 0.0022  &0.7035 &0.0040 \\ 
    LSTM-DNN  &  0.0940 &  0.5195 & 0.8335 & 0.1592 \\
    FCN-LSTM-DNN  & 0.4419  & 0.0023  & 0.0028 & 0.0045\\
    \textbf{ATSDLN}  & 0.3606  & 0.4512  &\textbf{0.4445} &\textbf{0.4010} \\
    \bottomrule
  \end{tabular}
\end{table}

\par
Table \ref{tab:t3} shows that, when the multi-detector detection result voting method is adopted, each detect needs to traverse all detectors and candidate run-time parameters combinations, it is very time-consuming, which do not meet the demands of industrial real-time monitoring scenarios. Moreover, compared to the voting algorithm, the neural network models behave higher F1 score, what's more, difference model structures can affect the evaluate metric. The CNN shows the better ability of abstract feature extraction in our task.
\subsection{Influence of share layers}
As previously reported, the model selects the run-time parameters for the current time series as well as the detector. The time series representation learned through the detector classification task will be used as the input of the run-time parameters selection task, which can assist the learning of the run-time parameters. This part discusses the influence of share layers. The relevant models are described as follows:
\begin{itemize}
\item {\verb|NS-Model|}: without shared of network.
\item {\verb|SSR-Model|}: shared the shallow representation (the output layer of FCN).
\item {\verb|ATSDLN|}: shared specific representation (the layers of the detector classification task).
\end{itemize}

\begin{table}
  \caption{Effect of sharing layers}
  \label{tab:t4}
  \begin{tabular}{lllll}
    \toprule
    \textbf{model type}    & \textbf{Precision} & \textbf{Recall} &\textbf{Error}&\textbf{F1} \\
    \midrule
    NS-Model  & 0.0982  & 0.5146  &0.8253 &0.1649 \\
    SSR-Model  & 0.2366  & 0.3374  &0.5212 &0.2782 \\
    \textbf{ATSDLN}  & 0.3606  & 0.4512  &\textbf{0.4445} &\textbf{0.4010} \\
    \bottomrule
  \end{tabular}
\end{table}
\par
As is shown in Table \ref{tab:t4}, the effect of shared the shallow and specific representations of time series is optimal through the classification network (anomaly detector selection) on the left and the classification network (run-time parameters selection) on the right. This is because the run-time parameters have a strong relativity with the detector. In order to output appropriate detector categories, the expression learned by the detector classification task will be used as input of the parameter classification task to assist parameter learning.
\subsection{Influence of Transfer learning}
To solve the shortage of data and let the network extract temporal features and initialize the models better, we selecting some UCR sample data sets for the comparative experiments of transfer learning.
\begin{itemize}
\item {\verb|ATSDLN|}: Training without transfer learning. 
\item {\verb|Transfer-1|}: Transfer from FordA to our data.
\item {\verb|Transfer-2|}: Transfer from Earthquakes to our data.
\item {\verb|Transfer-3|}: Transfer from coffe to our data.
\end{itemize}

\begin{table}
  \caption{ATSDLN with Transfer learning}
  \label{tab:t5}
  \begin{tabular}{lllll}
    \toprule
    \textbf{model type}    & \textbf{Precision} & \textbf{Recall} &\textbf{Error}&\textbf{F1} \\
    \midrule
    ATSDLN  & 0.3606  & 0.4512  &0.4445 &0.4010 \\
    \textbf{Transfer-1}  & 0.5191  & 0.3623  &\textbf{0.2513} &\textbf{0.4268} \\
    Transfer-2 & 0.3724  & 0.4350  &0.4230 &0.4013 \\
    Transfer-3 & 0.3613  & 0.4506  &0.4434 &0.4011\\
    \bottomrule
  \end{tabular}
\end{table}
\par
The second chapter mentions that transfer learning is applied to the FCN layer and fine-tuning in the FC layers, which makes the network parameters initialized better, so as to speed up the training and convergence and improve the performance of time series classification model. Table \ref{tab:t5} shows that, in most of the cases, the pre-trained model can improve the performance of model.
\section{Conclusions}
This paper proposed a new ATSDLN model, which realized an adaptional classification of time series anomaly detectors and run-time parameters selection by combining transfer learning and dynamic adaptive joint learning. The second Chapter mentions that the network is composed of two sub-tasks: anomaly detector classification and the run-time parameters fine-tuning network, both of which are supervised classification models. Because the size of the candidate run-time parameters sets of each detector is inconsistent, the last layer width of the right network (the run-time parameters fine-tuning network) follows the left as the side detector changes, that the model supports flexible addition and deletion detectors. Furthermore, because the run-time parameters have a strong relativity with the anomaly detector, the effect of shared the shallow and specific representations of time series is optimal through anomaly detectors classification network and run-time parameters fine-tuning network. Moreover, we pre-trained FCN layers based on different data sets, the results investigated that transfer learning approach can improve the performance of our model. Experiment results show that ATSDLN solves the problem of low precision and high false alarm ratio when the data pattern is change. ATSDLN is also applied to our industrial scenarios. In the future, we will consider extract global features of time series and alarm suppression.



\bibliographystyle{ACM-Reference-Format}
\bibliography{sample-base}

\appendix
\end{document}